\documentclass{article}
\usepackage{spconf,amsmath,graphicx,float}
\usepackage{hyperref}
\usepackage{amssymb}
\usepackage{bbding}
\usepackage{indentfirst}
\usepackage[dvipsnames]{xcolor}
\graphicspath{./images/}

\title{Multi-Granularity and Multi-modal Feature Interaction Approach for Text Video Retrieval}
\name{Wenjun Li$^{\ast}$, Shudong Wang$^{\ast}$, Dong Zhao$^{\ast}$, Shenghui Xu, Zhaoming Pan, Zhimin Zhang\thanks{$^{\ast}$Authors contributed equally to this work.}}
\address{Netease Media Group, Beijing, China}
\begin{document}
%
\maketitle
\begin{abstract}
The key of the text-to-video retrieval (TVR) task lies in learning the unique similarity between each pair of text (consisting of words) and video (consisting of audio and image frames) representations. However, some problems exist in the representation alignment of video and text, such as a text, and further each word, are of different importance for video frames. Besides, audio usually carries additional or critical information for TVR in the case that frames carry little valid information. Therefore, in TVR task, multi-granularity representation of text, including whole sentence and every word, and the modal of audio are salutary which are underutilized in most existing works. To address this, we propose a novel multi-granularity feature interaction module called MGFI, consisting of text-frame and word-frame, for video-text representations alignment.  Moreover, we introduce a cross-modal feature interaction module of audio and text called CMFI to solve the problem of insufficient expression of frames in the video. Experiments on benchmark datasets such as MSR-VTT, MSVD, DiDeMo show that the proposed method outperforms the existing state-of-the-art methods.
\end{abstract}
\section{Introduction and background}
\label{sec:intro}

Recently, as video has become the main way for people to enjoy entertainment and obtain information, the application of video on the Internet has experienced growth explosively. Faced with such a huge amount of videos, how to find similar videos accurately through text is becoming important increasingly, which directly affects the user experience on the video platform. Text Video Retrieval (TVR) uses text as query and calculates cosine similarity with video features to get the similarity ranking in descending order.

In recent years, language-image pre-trained models have rapid development, such as ViLBERT\cite{lu2019vilbert}, UNITER\cite{chen2020uniter}, CLIP\cite{radford2021learning}, ALIGN\cite{jia2021scaling}, WenLan\cite{huo2021wenlan}, HiVLP\cite{chen2022hivlp}, etc., which effectively connect the features of text and image, and show strong generalization and migration capabilities in many downstream applications. In the field of TVR, many excellent works based on pre-trained CLIP have emerged and achieved excellent performance. CLIP4Clip\cite{luo2021clip4clip} introduces no parameters by using a mean pooling mechanism on video frames, or introduce more parameters, e.g., the self-attention or transformer, to solve the text-video retrieval problem; CLIP2Video\cite{fang2021clip2video} proposes two independent modules, multi-modal learning of image-text and temporal relationships between video frames and video-text, to solve the multi-modal learning problems in spatial and temporal aspects respectively; CLIP2TV\cite{gao2021clip2tv} consists of video-text alignment module and video-text matching module, which can boost the performance to each other, and proposes similarity distillation to alleviate the impairment brought by data noise; CAMoE+DSL\cite{cheng2021improving} proposes a heterogeneity of structures called CAMoE to learn how to align cross-modal information and a novel Dual Softmax Loss (DSL) to revise the predicted similarity score; X-Pool\cite{gorti2022x} designs a text-conditioned video pooling and a cross-modal attention model to extract important visual cues according to text. 

However, through the observation and analysis of the datasets, the text and each word have different degrees of relevance for video frames in some cases, as some words are completely irrelevant and some words are closely related. Moreover, audio may have a good correlation to text when video frames carry little valid information. In this paper, we propose a new CLIP-based framework to incorporate text, words and audio into text-video retrieval task. The main contributions can be summarized as follows: 

a) We design a multi-granularity feature interaction module(MGFI) based on text-frame and word-frame to generate an aggregated video representation. 

b) We propose a cross-modal feature interaction module(CMFI) of text-audio for assisted correction, to enhance the contrastive learning between representations of video and text. 

c) In-depth and extensive experimental results on multiple datasets validate the effectiveness of the proposed method by conducting ablation studies on each module.

\section{proposed Methods}
\label{sec:methods}

In this section, we will introduce the overall architecture of our approach and elaborate on the submodules.

\begin{figure*}[ht] 
\centering 
\includegraphics[width=1.0\linewidth]{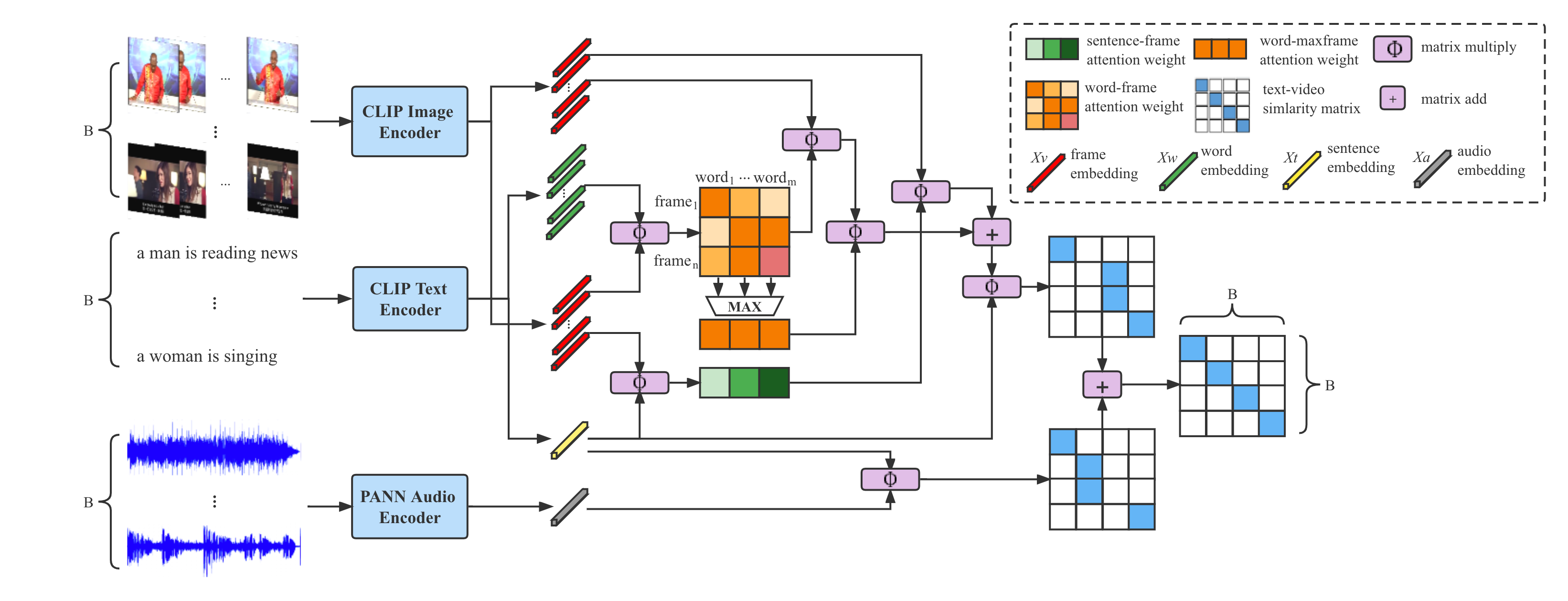} 
\caption{Framework of our approach which is comprised of two components: MGFI aligns video and text, and CMFI complements expression of frames information in video.} 
\label{fig1} 
\end{figure*}

\subsection{Overall Structure Design}
\label{ssec:overal}

The overall architecture shown in Figure \ref{fig1} has three input modalities: video frames, video caption and audio. Firstly, frames and text are encoded by pre-trained CLIP encoders and audio is encoded by PANN encoder which is pre-trained in a large-scale audio dataset. In this step, frame embedding, caption embedding including embedding of words and sentence, and audio embedding are obtained. Then, MGFI is performed on frame embedding, word embedding, and sentence embedding to calculate a video-text similarity matrix. Simultaneously, the audio-text similarity matrix is acquired by interacting with the sentence embedding and the audio embedding in CMFI module. Finally, the similarity matrices of video-text and audio-text are fused to perform contrastive learning between representations of video and text.

\subsection{Multi-Granularity Feature Interaction}
\label{ssec:cross}

This module is based on two fundamental observations. On the one hand, the content of video is much richer than text. The existing methods, averagely pooling the features of all frames in the video, will result in a huge sacrifice in performance. On the other hand, the importance of each word in the text for video retrieval is different. Likewise, if sentence feature is used to represent the entire text, the features of keywords may be overwhelmed by features of other words that are not semantically informative.

Based on these observations above, we propose the MGFI method for text-video retrieval, which combines frame-text and frame-word interaction. In the frame-text cross-modal interaction, different from the average pooling to obtain video representations adopted in existing methods, we obtain the video representation in a text-conditioned manner. Specifically, given a text representation, we obtain the importance of each frame in the video relative to the text representation through a cross-modal attention mechanism. Therefore, the video representation is obtained by weighted summation of features of all frames, where frames with higher semantic similarity to the text are more important to generate video representation. This process can be formally expressed as follows:
\begin{equation}
\small
    q=LN(x_t)W_q, k=LN(x_v)W_k, v=LN(x_v)W_v,
\end{equation}
\begin{equation}
\small
    z=\sigma(qk^\top / \sqrt{C})vW_o,
\end{equation}
\begin{equation}
\small
    o_1=z+FF(LN(z)),
\end{equation}
where $x_t\in \mathbb{R}^{C}, x_v \in \mathbb{R}^{N\times C}$ denotes the representation of the text and video, $C$ denotes the dimension and $N$ denotes the number of frames in the video. $LN$ denotes layer normalization. $FF$ denotes the feed-forward layer which is composed of two linear layers between which is a nonlinear activation layer. $W_q,W_k,W_v,W_o\in\mathbb{R}^{C \times C}$ denote the projection matrices. $z,q \in \mathbb{R}^{C}$, $k,v \in \mathbb{R}^{N \times C}$ denote the embedding output, query, key, value repectively. $\sigma(.)$ is the softmax function. $o_1 \in \mathbb{R}^{C}$ denotes the video representation corresponding to $x_t$.

For more fine-grained interactions, we further extend the frame-text interaction to frame-word interaction. Specifically, for each word in the text, a word-conditioned video representation is obtained in the similar manner of frame-text. Then we perform weighted fusion of these video representations to obtain a video representation relative to the entire text. For the video representation corresponding to each word, we take the maximum similarity between the word and each frame in the video as the measure of its weight:
\begin{equation}
\small
    \tilde{q}=LN(x_w)W_q,
\end{equation}
\begin{equation}
\small
    a=\tilde{q}k^\top,
\end{equation}
\begin{equation}
\small
    \tilde{z}=\sigma(a/\sqrt C)vW_o,
\end{equation}
\begin{equation}
\small
    \tilde{o}=\sigma(max(a))\tilde{z},
\end{equation}
\begin{equation}
\small
    o_2=\tilde{o}+FF(LN(\tilde{o})),
\end{equation}
where $x_w\ \in \mathbb{R}^{N' \times C}$ denotes the representation of the text at a word-granularity level, $N'$ denotes the number of words in the text. $\tilde{q} \in \mathbb{R}^{N' \times C}$ denotes the embedding query. $a \in \mathbb{R}^{N' \times N}$ denotes the similarity between each word and each frame. $\tilde{z} \in \mathbb{R}^{N' \times C}$ denotes embedding output corresponding to each word in the text. $\tilde{o} \in \mathbb{R}^{C}$ denotes the embedding output after weighted fusion. $o_2 \in \mathbb{R}^{C}$ denotes the video representation after word-frame fine-grained interactions.

Finally, $o_1$ and $o_2$ obtained by the two interactions are averaged to get the video representation $o \in \mathbb{R}^{C}$ and the video-text similarity matrix $s(v,t)$ is calculated as follow:
\begin{equation}
\small
    o=(o_1+o_2)/2,
\end{equation}
\begin{equation}
\small
    s(v,t) = \frac{o \cdot x_t}{\|o\|\|x_t\|},
\end{equation}

\subsection{Cross-Modal Feature Interaction}
Since audio plays an important role in the representation of most videos, especially when little valid information exists in frames, it is unreasonable to ignore the audio in text-video retrieval task. To fully utilize audio information, we introduce the CMFI module to improve the precision of text-video retrieval. Several feature interaction methods are implemented and verified, like fusion of the audio and video frames features through self-attention before modal interaction with text. By analyzing this result, we find that there is a lot of noise in audio and frames, and directly applying interaction in audio and video frames brings more distractions to video representation and leads to performance degradation. In contrast to video frames, text carries less but more explicit information, which is favorable for network to make use of valid feature of audio.

Considering the analysis above, in CMFI module, the audio-text similarity matrix is obtained through a product operation between the audio and text embeddings. We define our audio-text similarity function $s(a,t)$ as as below:

\begin{equation}
\small
    s(a,t) = \frac{L(LN(x_a)) \cdot x_t}{\|L(LN(x_a))\|\|x_t\|},
\end{equation}

where $x_a, x_t \in \mathbb{R}^{C}$ denotes the representation of audio and text, $C$ denotes the dimension. $LN$ denotes layer normalization. $L$ denotes a linear layer.

\subsection{Loss Function}
\label{ssec:interaction}

The Network is trained with InfoNCE\cite{oord2018representation} loss which is widely used in the field of contrastive learning. Given a batch of $B$, video-text pairs generate $B$ × $B$ similarity matrix, which can be optimized in the form of cross-entropy.
\begin{equation}
\small
    \mathcal Lv2t = -\frac{1}{B} \sum_{i}^B \log(\frac{\exp(s(v_i,t_i)+s(a_i,\tilde{t_i}))}{\sum_{j=1}^B \exp(s(v_i,t_j)+s(a_i,\tilde{t_j})})
\end{equation}
\begin{equation}
\small
    \mathcal Lt2v = -\frac{1}{B} \sum_{i}^B \log(\frac{\exp(s(v_i,t_i)+s(a_i,\tilde{t_i}))}{\sum_{j=1}^B \exp(s(v_j,t_i)+s(a_j,\tilde{t_i})})
\end{equation}

\begin{equation}
\small
    \mathcal L = \mathcal Lt2v + \mathcal Lv2t
\end{equation}
$s$ denotes the inner product of two embedding. $v_i$, $t_i$ represent the video and text representations after cross-modal interaction, respectively. $a_i$ represents the audio encoding result. $\tilde{t_i}$ denotes the original text embedding. The loss $\mathcal L$ is the sum of video-to-text loss $\mathcal Lv2t$ and text-to-video loss $\mathcal Lt2v$.

\section{experiments and results}
\label{sec:exp}
To validate the effectiveness of each module we proposed, we first conduct detailed ablation experiments on the MSR-VTT dataset\cite{xu2016msrvtt}, and then comprehensive experiments are conducted on the MSVD\cite{chen2011collecting}, MSR-VTT, DiDeMo\cite{anne2017localizing} datasets, respectively.

\subsection{Datasets And Metrics}
\label{ssec:dataset}
MSR-VTT consists of 10000 videos, each 10 to 32s in length and 20 items in cation. The training set is divided into 7K and 9K groups. The training set is one video corresponding to multiple captions, and the test set is one video corresponding to one caption. To validate our model, we use 9K as the training set and 1K as the test set for testing respectively.

MSVD is composed of 1970 videos with a split of 1200, 100, and 670 as the train, validation, and test set, respectively. Each video is paired with approximately 40 captions and ranges from 1 to 62 seconds.

DiDeMo contains 10,000 videos with 40,000 sentences. All captions are concatenated into a single query for text-video retrieval.

Evaluation Metric: We follow the standard retrieval task\cite{radford2021learning} and adopt Recall at rank K (R@K), median rank (MdR) and mean rank (MnR) as metrics. Higher R@K and lower MdR or MnR indicate better performance.


\subsection{Implementation Details}
\label{ssec:implement}

The video encoder and text encoder are both initialized by CLIP(ViT-B/32), and the audio encoder is initialized by PANN.
All embedding dimensions are set to 512, including frame, text and audio. We set our batch size to 32 for all experiments and optimize the model for 5 epochs by using AdamW. The learning rate for visual encoder and text encoder is 1e-6. As for audio encoder, we set a slightly larger learning rate to 5e-5. For the MSR-VTT and MSVD datasets, we sample 12 frames for each video and 30 frames for the DiDeMo dataset following previous works. We first train the network for video frames and text modal interaction. Then fix the network weights, and finetune the weights of the audio network.

\subsection{Ablation Study}
\label{ssec:ablaton}

We conduct detailed evaluation experiments for the MGFI, consisting of s-f and w-f, and CMFI, consisting of a-s, on the MSR-VTT-9K dataset. The results of the ablation experiments are shown in Table \ref{tabel:ablation-study}. The t2v means retrieving video by text.

\begin{table}[ht]
\small
\begin{center}
\renewcommand{\arraystretch}{1.1}
\begin{tabular}{cccc|ccc}
    \hline
    base&s-f&w-f&a-s&R@1&R@5&R@10\\
    \hline
    \Checkmark & & & & 43.1 & 70.4 &  80.8\\
    \Checkmark & \Checkmark &   &   & 46.8 & 72.2 & 82.8\\
    \Checkmark &  & \Checkmark &   & 46.7 & 72.3 & 83.0\\
    \Checkmark & \Checkmark & \Checkmark &   & 47.1 & 72.8 & 82.6\\
    \Checkmark &  & \Checkmark & \Checkmark & 46.9 & 74.0 & 83.2\\
    \Checkmark & \Checkmark &   & \Checkmark & 48.2 & 73.0 & 82.2\\
    \Checkmark & \Checkmark & \Checkmark & \Checkmark & 48.4 & 73.1 & 83.6\\
    \hline
\end{tabular}
\caption{ablation study of t2v result on feature interaction}
\label{tabel:ablation-study}
\end{center}
\end{table}

In Table \ref{tabel:ablation-study}, the base model is CLIP4Clip without modal interaction, s-f, w-f and a-s means that feature interaction between sentences and frames,  words and frames, audio and sentences respectively. It can be observed that s-f module, w-f module, and a-f module bring about the absolute improvement of 3.7\%, 3.6\%, 1.4\% in R@1. Through in-depth analysis, s-f module can strengthen the matching degree between each sentence and the key frame, w-f module further strengthens the matching of detail granularity, and a-s module supplies the missing information in the matching process between frames and texts, so that the index of retrieval can be improved. We conducted the permutation and combination experiments between each two modules at the same time. It can be observed that the improvement of indicators brought by each module is not mutually exclusive, but complements each other. Finally, by combining s-f, w-f, and a-s, an R@1 of 48.4 was obtained, and the index increased by 5.3\%, reaching the state-of-the-art level.

\subsection{Comparison With State-of-the-arts}
\label{ssec:comparison}

In this section, we compare the results of proposed method with some state-of-the-art methods on MSVD, MSR-VTT, and DiDeMo datasets. All of these methods are based on the CLIP pre-trained model as the backbone, and the image encoder is ViT-B/32.

It can be seen in Table \ref{table:msr-vtt-9k} that our method outperforms other methods on the MSR-VTT-9K dataset. It surpassed CLIP4Clip by 12.2\% on the relative value indicator of R@1, reaching the current state-of-the-art. 

\begin{table}[ht]
\small
\begin{center}
\renewcommand{\arraystretch}{1.1}
\begin{tabular}{c|ccccc}
 \hline
 Methods & R@1 & R@5 & R@10 & MdR & MnR \\ 
 \hline
 CLIP\cite{radford2021learning} & 31.2 & 53.7 & 64.2 & 4.0 & - \\ 
 CLIP4Clip\cite{luo2021clip4clip} & 43.1 & 70.4 &  80.8 & 2.0 & 16.2 \\ 
 CLIP2Video\cite{fang2021clip2video} & 45.6 & 72.6 & 81.7 & 2.0 & 14.6 \\
 CLIP2TV+SD\cite{gao2021clip2tv} & 46.1 & 72.5 & 82.9 & 2.0 & 15.2 \\
 CAMoE+DSL\cite{cheng2021improving} & 47.3 & 74.2 & 84.5 & 2.0 & 11.9 \\ 
 X-Pool\cite{gorti2022x} & 47.2 & 72.8 & 82.6 & 2.0 & 13.8 \\
 \hline
 \textbf{Ours} & \textbf{48.4} & 73.1 & 83.6 & 2.0 & 13.3 \\
 \hline
\end{tabular}
\caption{t2v results on the MSR-VTT-9K dataset}
\label{table:msr-vtt-9k}
\end{center}
\end{table}

\begin{table}[ht]
\small
\begin{center}
\renewcommand{\arraystretch}{1.1}
\begin{tabular}{c|ccccc} 
 \hline
 Methods & R@1 & R@5 & R@10 & MdR & MnR \\ 
 \hline
 CLIP\cite{radford2021learning} & 37.0 & 64.1 & 73.8 & 3.0 & - \\ 
 CLIP4Clip\cite{luo2021clip4clip} & 46.2 & 76.1 & 84.6 & 2.0 & 10.0 \\
 CLIP2Video\cite{fang2021clip2video} & 47.0 & 76.8 & 85.9 & 2.0 & 9.6 \\
 CLIP2TV+SD\cite{gao2021clip2tv} & 47.0 & 76.5 & 85.1 & 2.0 & 10.1 \\
 CAMoE+DSL\cite{cheng2021improving} & 49.8 & 79.2 & 87.0 & 2.0 & 9.4 \\
 X-Pool\cite{gorti2022x} & 47.2 & 77.4 & 86.0 & 2.0 & 9.3 \\
 \hline
 \textbf{Ours} & \textbf{53.5} & \textbf{81.0} & \textbf{88.2} & \textbf{1.0} & \textbf{7.7}\\
 \hline
\end{tabular}
\caption{t2v results on the MSVD dataset}
\label{table:msvd}
\end{center}
\end{table}

On the other two datasets, MSVD and DiDeMo, it also reaches state-of-the-art level, shown in Table \ref{table:msvd} \ref{table:didemo}. The proposed method improves the relative value of CLIP4Clip R@1 by 15.8\% on the MSVD and achieves remarkable performance 46.1\% R@1 with a relative performance improvement of 5.2\% compared to CLIP4Clip for t2v on DiDeMo.

\begin{table}[ht]
\small
\begin{center}
\renewcommand{\arraystretch}{1.1}
\begin{tabular}{c| c c c c c} 
 \hline
 Methods & R@1 & R@5 & R@10 & MdR & MnR \\
 \hline
 ClipBERT\cite{lei2021less} & 20.4 & 48.0 & 60.8 & 6.0 & - \\ 
 CLIP4Clip\cite{luo2021clip4clip} & 43.4 & 70.2 & 80.6 & 2.0 & 17.5 \\ 
 CLIP2TV+SD\cite{gao2021clip2tv} & 45.5 & 69.7 & 80.6 & 2.0 & 17.1 \\
 CAMoE+DSL\cite{cheng2021improving} & 43.8 & 71.4 & - & - & - \\
 \hline
 \textbf{Ours} & \textbf{46.1} & \textbf{73.0} & \textbf{82.6} & \textbf{2.0} & \textbf{13.7} \\
 \hline
\end{tabular}
\caption{t2v results on the DiDeMo dataset}
\label{table:didemo}
\end{center}
\end{table}

\section{conclusion}
\label{sec:conclusion}
To solve the problem that the alignment of video frames and words is neglected and audio information of video is not fully utilized, we design a MGFI module based on text-frame and word-frame to generate an aggregated video representation, and propose a CMFI module of text-audio to enhance the contrastive learning between representations of video and text. Experimental results show that using the cooperative relationship among sentence-frame, word-frame, and audio-sentence multiple modalities in TVR is actually meaningful and can greatly improve information utilization and retrieval accuracy.


\vfill\pagebreak

\bibliographystyle{IEEEbib}
\bibliography{refs_strings,refs}

\begin{thebibliography}{10}

\bibitem{lu2019vilbert}
Jiasen Lu, Dhruv Batra, Devi Parikh, and Stefan Lee,
\newblock ``Vilbert: Pretraining task-agnostic visiolinguistic representations
  for vision-and-language tasks,''
\newblock {\em Advances in neural information processing systems}, vol. 32,
  2019.

\bibitem{chen2020uniter}
Yen-Chun Chen, Linjie Li, Licheng Yu, Ahmed El~Kholy, Faisal Ahmed, Zhe Gan,
  Yu~Cheng, and Jingjing Liu,
\newblock ``Uniter: Universal image-text representation learning,''
\newblock in {\em European conference on computer vision}. Springer, 2020, pp.
  104--120.

\bibitem{radford2021learning}
Alec Radford, Jong~Wook Kim, Chris Hallacy, Aditya Ramesh, Gabriel Goh,
  Sandhini Agarwal, Girish Sastry, Amanda Askell, Pamela Mishkin, Jack Clark,
  et~al.,
\newblock ``Learning transferable visual models from natural language
  supervision,''
\newblock in {\em International Conference on Machine Learning}. PMLR, 2021,
  pp. 8748--8763.

\bibitem{jia2021scaling}
Chao Jia, Yinfei Yang, Ye~Xia, Yi-Ting Chen, Zarana Parekh, Hieu Pham, Quoc Le,
  Yun-Hsuan Sung, Zhen Li, and Tom Duerig,
\newblock ``Scaling up visual and vision-language representation learning with
  noisy text supervision,''
\newblock in {\em International Conference on Machine Learning}. PMLR, 2021,
  pp. 4904--4916.

\bibitem{huo2021wenlan}
Yuqi Huo, Manli Zhang, Guangzhen Liu, Haoyu Lu, Yizhao Gao, Guoxing Yang,
  Jingyuan Wen, Heng Zhang, Baogui Xu, Weihao Zheng, et~al.,
\newblock ``Wenlan: Bridging vision and language by large-scale multi-modal
  pre-training,''
\newblock {\em arXiv preprint arXiv:2103.06561}, 2021.

\bibitem{chen2022hivlp}
Feilong Chen, Xiuyi Chen, Jiaxin Shi, Duzhen Zhang, Jianlong Chang, and
  Qi~Tian,
\newblock ``Hivlp: Hierarchical vision-language pre-training for fast
  image-text retrieval,''
\newblock {\em arXiv preprint arXiv:2205.12105}, 2022.

\bibitem{luo2021clip4clip}
Huaishao Luo, Lei Ji, Ming Zhong, Yang Chen, Wen Lei, Nan Duan, and Tianrui Li,
\newblock ``Clip4clip: An empirical study of clip for end to end video clip
  retrieval,''
\newblock {\em arXiv preprint arXiv:2104.08860}, 2021.

\bibitem{fang2021clip2video}
Han Fang, Pengfei Xiong, Luhui Xu, and Yu~Chen,
\newblock ``Clip2video: Mastering video-text retrieval via image clip,''
\newblock {\em arXiv preprint arXiv:2106.11097}, 2021.

\bibitem{gao2021clip2tv}
Zijian Gao, Jingyu Liu, Sheng Chen, Dedan Chang, Hao Zhang, and Jinwei Yuan,
\newblock ``Clip2tv: An empirical study on transformer-based methods for
  video-text retrieval,''
\newblock {\em arXiv preprint arXiv:2111.05610}, 2021.

\bibitem{cheng2021improving}
Xing Cheng, Hezheng Lin, Xiangyu Wu, Fan Yang, and Dong Shen,
\newblock ``Improving video-text retrieval by multi-stream corpus alignment and
  dual softmax loss,''
\newblock {\em arXiv preprint arXiv:2109.04290}, 2021.

\bibitem{gorti2022x}
Satya~Krishna Gorti, No{\"e}l Vouitsis, Junwei Ma, Keyvan Golestan, Maksims
  Volkovs, Animesh Garg, and Guangwei Yu,
\newblock ``X-pool: Cross-modal language-video attention for text-video
  retrieval,''
\newblock in {\em Proceedings of the IEEE/CVF Conference on Computer Vision and
  Pattern Recognition}, 2022, pp. 5006--5015.

\bibitem{oord2018representation}
Aaron van~den Oord, Yazhe Li, and Oriol Vinyals,
\newblock ``Representation learning with contrastive predictive coding,''
\newblock {\em arXiv preprint arXiv:1807.03748}, 2018.

\bibitem{xu2016msrvtt}
Jun Xu, Tao Mei, Ting Yao, and Yong Rui,
\newblock ``Msr-vtt: A large video description dataset for bridging video and
  language,''
\newblock in {\em IEEE International Conference on Computer Vision and Pattern
  Recognition (CVPR)}, June 2016.

\bibitem{chen2011collecting}
David Chen and William~B Dolan,
\newblock ``Collecting highly parallel data for paraphrase evaluation,''
\newblock in {\em Proceedings of the 49th annual meeting of the association for
  computational linguistics: human language technologies}, 2011, pp. 190--200.

\bibitem{anne2017localizing}
Lisa Anne~Hendricks, Oliver Wang, Eli Shechtman, Josef Sivic, Trevor Darrell,
  and Bryan Russell,
\newblock ``Localizing moments in video with natural language,''
\newblock in {\em Proceedings of the IEEE international conference on computer
  vision}, 2017, pp. 5803--5812.

\bibitem{lei2021less}
Jie Lei, Linjie Li, Luowei Zhou, Zhe Gan, Tamara~L Berg, Mohit Bansal, and
  Jingjing Liu,
\newblock ``Less is more: Clipbert for video-and-language learning via sparse
  sampling,''
\newblock in {\em Proceedings of the IEEE/CVF Conference on Computer Vision and
  Pattern Recognition}, 2021, pp. 7331--7341.

\end{thebibliography}

\end{document}